\documentclass[letterpaper, 10 pt, conference]{ieeeconf}  

\IEEEoverridecommandlockouts                              

\usepackage{graphicx}
\usepackage{cite}
\usepackage{picinpar}
\usepackage{amsmath}
\usepackage{url}
\usepackage{flushend}
\usepackage[latin1]{inputenc}
\usepackage{colortbl}
\usepackage{soul}
\usepackage{multirow}
\usepackage{pifont}
\usepackage{color}
\usepackage{alltt}
\usepackage{enumerate}
\usepackage{siunitx}
\usepackage{epstopdf}
\usepackage{pbox}
\usepackage{amssymb}
\makeatletter
\let\NAT@parse\undefined
\makeatother
\usepackage{hyperref}
\let\labelindent\relax
\usepackage{enumitem}
\usepackage{relsize}
\usepackage{booktabs}
\usepackage{xcolor}

\title{\LARGE \bf
MoCap2GT: A High-Precision Ground Truth Estimator for SLAM Benchmarking Based on Motion Capture and IMU Fusion
}

\author{Zichao Shu$^{1,2}$, Shitao Bei$^{2}$, Jicheng Dai$^{2}$, Lijun Li$^{2}$, and Zetao Chen$^{2}$
\thanks{$^{1}$Shanghai Institute of Technical Physics, Chinese Academy of Sciences, Shanghai, China. \tt\footnotesize zichao-shu@ylab.ac.cn}
\thanks{$^{2}$Advanced Display and Sensing Research Center, Yongjiang Laboratory, Ningbo, China. \tt\footnotesize\{shitao-bei; jicheng-dai; lijun-li; zetao-chen\}@ylab.ac.cn}
}

\begin{document}

\maketitle
\thispagestyle{empty}
\pagestyle{empty}

\begin{abstract}

Marker-based optical motion capture (MoCap) systems are widely used to provide ground truth (GT) trajectories for benchmarking SLAM algorithms. However, the accuracy of MoCap-based GT trajectories is mainly affected by two factors: spatiotemporal calibration errors between the MoCap system and the device under test (DUT), and inherent MoCap jitter. Consequently, existing benchmarks focus primarily on absolute translation error, as accurate assessment of rotation and inter-frame errors remains challenging, hindering thorough SLAM evaluation. This paper proposes MoCap2GT, a joint optimization approach that integrates MoCap data and inertial measurement unit (IMU) measurements from the DUT for generating high-precision GT trajectories. MoCap2GT includes a robust state initializer to ensure global convergence, introduces a higher-order B-spline pose parameterization on the $SE(3)$ manifold with variable time offset to effectively model MoCap factors, and employs a degeneracy-aware measurement rejection strategy to enhance estimation accuracy. Experimental results demonstrate that MoCap2GT outperforms existing methods and significantly contributes to precise SLAM benchmarking. The source code is available at \url{https://anonymous.4open.science/r/mocap2gt} (temporarily hosted anonymously for double-blind review).

\end{abstract}

\setlength{\baselineskip}{0.92\baselineskip}

\section{Introduction} \label{SEC: Introduction}

Simultaneous localization and mapping (SLAM) enables egomotion estimation and geometric reconstruction, leading to extensive applications in various domains such as robotics, extended reality (XR), and autonomous driving. In recent years, visual-inertial (VI) SLAM algorithms have gained significant attention and experienced rapid development \cite{peng2024dvi, fan2024schurvins, campos2021orb}. Performance benchmarking is essential in VI-SLAM research, as it involves comparing estimated trajectories with high-precision ground truth (GT) data. Various metrics are employed for evaluation, and the most commonly used are absolute translation and rotation errors (ATE/ARE), which reflect global trajectory accuracy, and relative translation and rotation errors (RTE/RRE), which indicate errors over a short period \cite{sturm2012benchmark, jinyu2019survey}. Many recent VI-SLAM algorithms \cite{xu2024d, guo2024cmax, katragadda2024nerf} and commonly used benchmark datasets \cite{burri2016euroc, schubert2018tum, mueggler2017event} rely on motion capture (MoCap) systems, such as Vicon and OptiTrack, to provide GT trajectories within a controlled space. These systems adopt an outside-in localization paradigm, achieving drift-free six-degree-of-freedom (6-DoF) pose estimation by tracking reflective markers that are rigidly mounted on the device under test (DUT).

Nevertheless, the precision of MoCap-based GT trajectories is limited by two primary problems. First, the MoCap system and the DUT operate with independent local clocks and are defined in different reference frames. Since VI-SLAM typically adopts the inertial measurement unit (IMU) frame as its body frame, MoCap measurements must be temporally and spatially aligned with the built-in IMU of the DUT. However, this calibration process is subject to errors. Existing methods may introduce ATE and ARE in the resulting GT trajectories, which can exceed 2 mm and 0.3$^\circ$, respectively \cite{pedrosa2021general, furrer2018evaluation}. Second, MoCap is prone to measurement errors, mainly caused by noise in tracking infrared markers. Although these errors do not accumulate over time, they appear as high-frequency jitter at the same rate as the MoCap output. While often overlooked in many studies, such jitter can introduce inter-frame RTE and RRE of up to 0.5 mm and 0.1$^\circ$, respectively \cite{burri2016euroc, schubert2018tum}. It also compounds with calibration errors, further increasing the absolute error of the GT trajectories. Some studies have attempted to compensate for this jitter using IMU measurements \cite{burri2016euroc, geneva2020vicon2gt}, but issues such as limited accuracy and degradation remain.

\subsection{Motivation} \label{SEC: Motivation}

The accuracy of SLAM algorithms is continually increasing. In standard indoor scenarios, state-of-the-art (SOTA) SLAM can achieve ATE/ARE below 10 mm/1$^\circ$ and inter-frame RTE/RRE below 2 mm/0.1$^\circ$ at camera frame rates \cite{jinyu2019survey, campos2021orb, xu2025airslam}. This progress presents new challenges for obtaining GT with even higher precision for reliable benchmarking. MoCap-based GT estimated by existing methods enables accurate evaluation of ATE but struggles to precisely quantify rotation and inter-frame errors. Consequently, most current benchmarks primarily focus on ATE \cite{campos2021orb, fan2024schurvins, xu2025airslam}, despite the fact that other error metrics are equally important for a comprehensive assessment of SLAM performance. In this work, we aim to estimate high-precision GT trajectories based on MoCap measurements to meet the benchmarking requirements of SOTA SLAM. Specifically, the target accuracy is ATE/ARE $<$ 2 mm/0.2$^\circ$ and RTE/RRE $<$ 0.4 mm/0.02$^\circ$---an error level half an order of magnitude lower than that of SLAM algorithms. 

\subsection{Contributions}

We propose MoCap2GT, a method for estimating GT trajectories to support the benchmarking of SOTA SLAM algorithms. The approach fuses MoCap data with IMU measurements from the DUT and performs batch estimation within a maximum likelihood estimation (MLE) framework. Effectively, it achieves spatiotemporal calibration and leverages the short-term robustness of IMU measurements to compensate for high-frequency jitter in the MoCap data. To the best of our knowledge, this is the first approach that fulfills the accuracy targets specified in Section \ref{SEC: Motivation} using a consumer-grade IMU and MoCap system (details in Table~\ref{TAB: Comparison on Real-World Datasets} later). Our main contributions are as follows:

\begin{enumerate}[label=\arabic*.]
    \item We propose a GT estimator that begins with a robust state initializer, followed by joint optimization of MoCap and IMU data. It simultaneously performs spatiotemporal calibration and leverages the complementary strengths of the two sensors to mitigate errors, producing a high-precision, IMU-centric GT trajectory.
    \item A cubic B-spline on the $SE(3)$ manifold with a variable time offset is designed to effectively model MoCap factors, and degenerate measurements are identified to improve estimation accuracy.
    \item The proposed method is validated on both simulated and real-world datasets, demonstrating superior performance and practical value for SLAM benchmarking. The implementation is released as open-source to benefit the community.
\end{enumerate}

\section{Related Works} \label{SEC: Related Works}

This section revisits the two main problems in MoCap-based GT estimation---spatiotemporal calibration and jitter mitigation---and motivates the design adopted in our work.

\subsection{Spatiotemporal Calibration}
In many SLAM studies and datasets that rely on MoCap for GT, the focus primarily lies on spatiotemporal calibration \cite{jinyu2019survey, schubert2018tum, gao2022vector, dong2024sevar}. This is typically achieved through a two-stage method. First, the calibration between the MoCap system and the camera on the DUT is performed, commonly modeled as a classic hand-eye calibration problem, with the MoCap serving as the `hand' and the camera as the `eye'. The process involves calculating the time offset and estimating the homogeneous transformation between them \cite{wu2020correspondence}. During calibration, the MoCap outputs the trajectory of the markers while the camera captures the calibration patterns. This enables independent camera trajectory estimation and hand-eye calibration via a pose-to-pose loosely coupled method \cite{wu2021simultaneous}. Alternatively, calibration can also be achieved through a tightly coupled approach using constraints based on camera reprojection errors, which can provide higher accuracy \cite{pedrosa2021general, gao2022vector}. In the second stage, calibration is performed between the camera and IMU \cite{furgale2012continuous, yang2024multi}, and the results are propagated. The requirement for two separate calibrations leads to error accumulation, limiting the precision of these methods. It is more effective to directly calibrate between the MoCap and IMU. In some studies, spatiotemporal calibration is achieved by processing the trajectories from the IMU-centric SLAM and the MoCap \cite{furrer2018evaluation, shu2024spatiotemporal}. However, these methods are constrained by inherent precision of the SLAM.

\subsection{MoCap Jitter Mitigation}
MoCap jitter is typically regarded as a minor error and has not received sufficient attention. However, as SLAM benchmarking demands grow increasingly stringent, the impact of jitter becomes more significant, especially on relative error evaluation. Some efforts attempt to directly smooth MoCap data, such as the median filtering applied in the TUM-VI dataset \cite{schubert2018tum}. This reduces jitter but offers limited improvement in accuracy. More effective approaches jointly optimize MoCap and the DUT's built-in IMU within an MLE framework, achieving both spatiotemporal calibration and jitter mitigation. The well-known EuRoC MAV dataset \cite{burri2016euroc} employed a batch estimator that fuses MoCap and IMU data to estimate high-precision GT trajectories. However, this estimator is sensitive to motion degeneracy, and some sequences have been shown to suffer from inaccurate calibration \cite{geneva2020openvins, shu2024spatiotemporal}. Another method, Vicon2GT \cite{geneva2020vicon2gt}, builds optimization residuals from IMU preintegration and linear interpolation of MoCap data. It has been widely used for SLAM benchmarking \cite{zhang2022dido, katragadda2024nerf}. However, due to the lack of an initialization module, it cannot guarantee global convergence. Moreover, the linear interpolation representation of MoCap data makes it sensitive to noise. General multisensor calibration methods based on continuous-time motion representation can also be extended to MoCap-IMU fusion  \cite{rehder2016general, chen2025ikalibr}. For example, the widely used Kalibr framework \cite{rehder2016extending} includes a MoCap branch for GT trajectory estimation.

While these methods provide a basis for high-precision GT estimation, two main issues remain. First, they assume a constant time offset between sensors. In practice, MoCap and IMU operate on independent clocks without communication, and clock scale error can cause the offset to drift. For consumer-grade IMUs, this drift can exceed 2 ms per minute \cite{mueggler2017event}, requiring a variable time alignment scheme. Second, detecting low-excitation segments in sensor measurements and mitigating their impact on estimation is required.

\section{Notation and Problem Statement} \label{SEC: Notation and Problem Statement}

\begin{figure}[t]
\centering
\includegraphics[width=\linewidth]{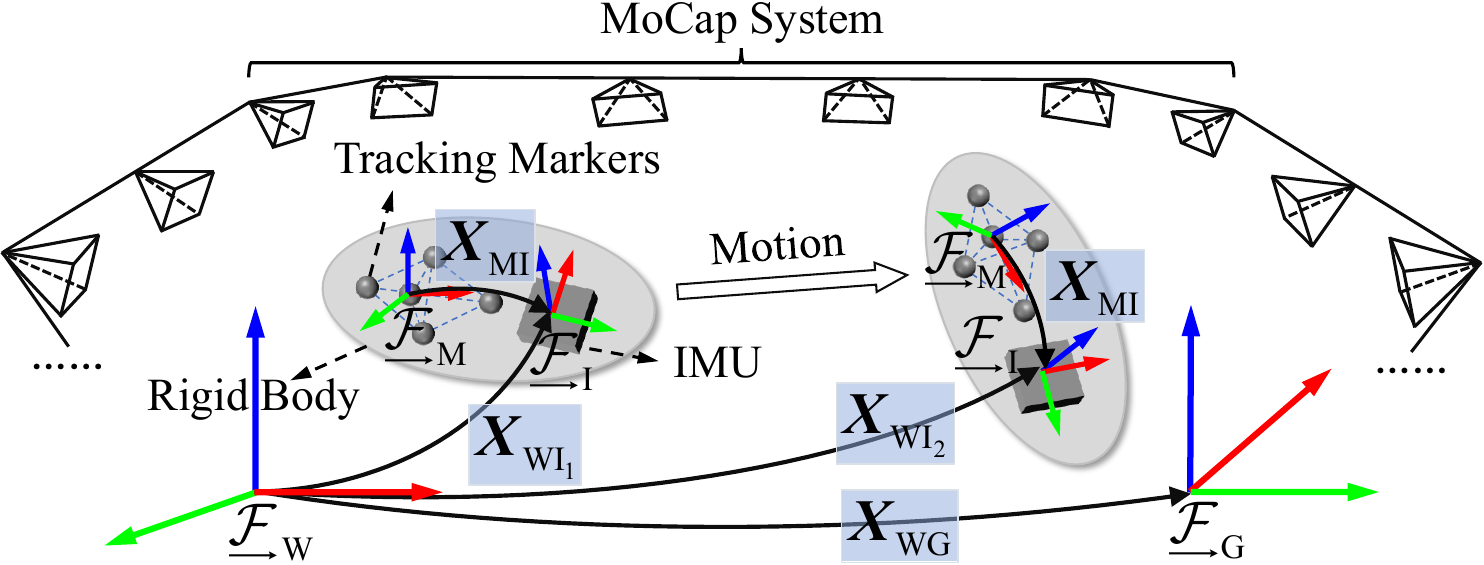}
\caption{Conventions for sensor coordinate frames and the states to be estimated (denoted as \textit{X}) in the proposed system.}
\label{FIG: Conventions for sensor frames}
\vspace{-5pt}
\end{figure}

In this section, we present the notation used throughout the paper and provide an overview of the problems we aim to address. A concrete illustration is shown in Fig. \ref{FIG: Conventions for sensor frames}.

\subsection{Notation}

\begin{enumerate}[label=\arabic*., itemsep=0.3ex]
    \item $\underrightarrow{\mathcal{F}}_{\mathrm{W}}$: The world coordinate frame of the MoCap system.
    \item $\underrightarrow{\mathcal{F}}_{\mathrm{M}}$: The marker-centric body frame of the MoCap.
    \item $\underrightarrow{\mathcal{F}}_{\mathrm{G}}$: The gravity-aligned world frame of the IMU.
    \item $\underrightarrow{\mathcal{F}}_{\mathrm{I}}$: The body frame of the IMU.
    \item $\boldsymbol{X}$: State vectors to be estimated in the proposed system, with specific meanings defined by the subscripts.
    \item $\boldsymbol{T}_{\scriptscriptstyle\mathrm{AB}} \in SE\left(3\right)$: The homogeneous transformation from frame $\underrightarrow{\mathcal{F}}_{\mathrm{B}}$ to frame $\underrightarrow{\mathcal{F}}_{\mathrm{A}}$, consisting of a rotation matrix $\boldsymbol{R}_{\scriptscriptstyle\mathrm{AB}} \in SO\left(3\right)$ and a translation vector $\boldsymbol{t}_{\scriptscriptstyle\mathrm{AB}} \in \mathbb{R}^3$. $\boldsymbol{R}_{\scriptscriptstyle\mathrm{AB}}$ can also be represented by a Hamilton quaternion $\boldsymbol{q}_{\scriptscriptstyle\mathrm{AB}}$.
    \item $\otimes$: Multiplication of Hamilton quaternions.
\end{enumerate}

\subsection{Problem Statement}

The estimator aims to spatiotemporally calibrate the MoCap and IMU and estimate the GT trajectory consistent with the IMU body frame $\underrightarrow{\mathcal{F}}_{\mathrm{I}}$ and clock, for SLAM benchmarking. To simplify the MoCap measurement model, the estimated trajectory is defined in the MoCap world frame $\underrightarrow{\mathcal{F}}_{\mathrm{W}}$. At each timestamp, the pose in the GT trajectory is represented as $\boldsymbol{T}_{\scriptscriptstyle\mathrm{WI}}$. The system relies on the 6-DoF pose measurement $\tilde{\boldsymbol{T}}_{\scriptscriptstyle\mathrm{WM}}$ from the MoCap and the 6-DoF inertial measurements $\tilde{\boldsymbol{a}}_{\scriptscriptstyle\mathrm{I}}$ and $\tilde{\boldsymbol{w}}_{\scriptscriptstyle\mathrm{I}}$ from the IMU. To enable joint optimization of the two sensors, we estimate the spatial extrinsic parameters $\boldsymbol{T}_{\scriptscriptstyle\mathrm{MI}}$ between body frames, the time offset $t_{\scriptscriptstyle\mathrm{MI}}$, and the rotation for aligning the world frames. Additionally, the velocity $\boldsymbol{v}{\scriptscriptstyle\mathrm{WI}}$, the accelerometer bias $\boldsymbol{b}{\scriptscriptstyle{a}}$, and the gyroscope bias $\boldsymbol{b}_{\scriptscriptstyle{g}}$ are estimated to enable IMU propagation. In summary, we jointly estimate the following states (indicated by solid black arrows in Fig. \ref{FIG: Conventions for sensor frames}):
\begin{align}
\boldsymbol{X} &= \big[ \boldsymbol{X}_{\scriptscriptstyle\mathrm{WI}_1}^\mathrm{\scriptscriptstyle{T}}, \dots, \boldsymbol{X}_{\scriptscriptstyle\mathrm{WI}_n}^\mathrm{\scriptscriptstyle{T}}, \boldsymbol{X}_{\scriptscriptstyle\mathrm{MI}}^\mathrm{\scriptscriptstyle{T}}, \boldsymbol{X}_{\scriptscriptstyle\mathrm{WG}}^\mathrm{\scriptscriptstyle{T}} \big]^\mathrm{\scriptscriptstyle{T}}, \label{EQ: All states} \\
\boldsymbol{X}_{\scriptscriptstyle\mathrm{WI}_k} &= \big[ \boldsymbol{p}_{\scriptscriptstyle\mathrm{WI}_k}^\mathrm{\scriptscriptstyle{T}}, \boldsymbol{v}_{\scriptscriptstyle\mathrm{WI}_k}^\mathrm{\scriptscriptstyle{T}}, \boldsymbol{q}_{\scriptscriptstyle\mathrm{WI}_k}^\mathrm{\scriptscriptstyle{T}}, \boldsymbol{b}_{\scriptscriptstyle{a, \mathrm{I}_k}}^\mathrm{\scriptscriptstyle{T}}, \boldsymbol{b}_{\scriptscriptstyle{g, \mathrm{I}_k}}^\mathrm{\scriptscriptstyle{T}} \big]^\mathrm{\scriptscriptstyle{T}}, \label{EQ: Inertial states} \\
\boldsymbol{X}_{\scriptscriptstyle\mathrm{MI}} &= \big[ \boldsymbol{p}_{\scriptscriptstyle\mathrm{MI}}^\mathrm{\scriptscriptstyle{T}}, \boldsymbol{q}_{\scriptscriptstyle\mathrm{MI}}^\mathrm{\scriptscriptstyle{T}}, t_{\scriptscriptstyle\mathrm{MI}_1}, \dots, t_{\scriptscriptstyle\mathrm{MI}_m} \big]^\mathrm{\scriptscriptstyle{T}}, \label{EQ: Spatiotemporal extrinsic states} \\
\boldsymbol{X}_{\scriptscriptstyle\mathrm{WG}} &= \big[ \theta_{\scriptscriptstyle\mathrm{WG}}^{\scriptscriptstyle\mathrm{roll}}, \theta_{\scriptscriptstyle\mathrm{WG}}^{\scriptscriptstyle\mathrm{pitch}} \big]^\mathrm{\scriptscriptstyle{T}}, \label{EQ: Gravity alignment states}
\end{align}
where $n$ inertial states are estimated at a specified frequency. To enable variable time offset estimation, we introduce $m$ time alignment control points. A MoCap timestamp $\tau_{\scriptscriptstyle\mathrm{M}}$ is converted to the IMU clock as $t_{\scriptscriptstyle\mathrm{M}} = \tau_{\scriptscriptstyle\mathrm{M}} - t_{\scriptscriptstyle\mathrm{MI}}$. Since the yaw angle is unobservable in gravity alignment, it is set to zero, and only the 2-DoF roll-pitch rotation between the world frames is estimated.

\section{Methodology} \label{SEC: Methodology}

\begin{figure}[t]
\centering
\includegraphics[width=\linewidth]{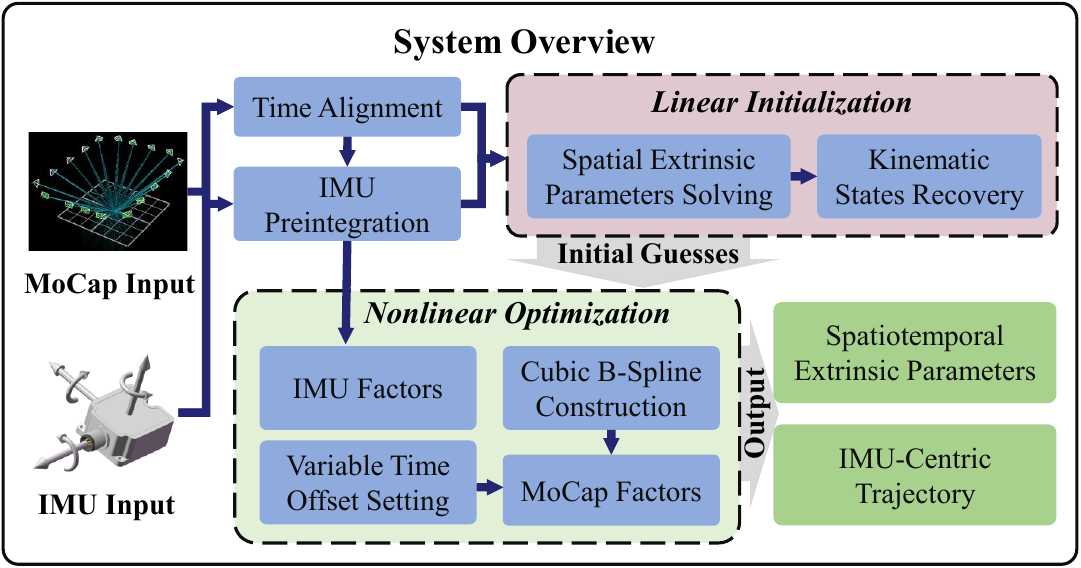}
\caption{System overview of the proposed estimator. The algorithm is composed of two main submodules: initialization and optimization, detailed in Sections \ref{SEC: Linear Initialization} and \ref{SEC: Nonlinear Optimization}, respectively. }
\label{FIG: System overview}
\vspace{-5pt}
\end{figure}

To estimate the target states, a coarse-to-fine estimator is proposed, as shown in Fig. \ref{FIG: System overview}. It first performs linear initialization without prior information to promote global convergence, followed by nonlinear refinement. Details of both modules are provided below.

\subsection{Linear Initialization} \label{SEC: Linear Initialization}

\subsubsection{IMU Preintegration} Initialization begins with a coarse time alignment to account for clock differences between sensors. With an initial assumption of a constant time offset between the MoCap and IMU, we apply a correlation method \cite{shu2024spatiotemporal} based on angular velocity signals. Once aligned, spatial calibration is achieved using rigid body motion constraints. Given two IMU timestamps $t_{\mathrm{I}_{\scriptscriptstyle{i}}}$ and $t_{\mathrm{I}_{\scriptscriptstyle{j}}}$, IMU data are integrated to propagate states and expressed in preintegration form with $\underrightarrow{\mathcal{F}}_{\mathrm{I}_{\scriptscriptstyle{i}}}$ as the reference frame: \par\vspace{-3pt}
\begin{footnotesize}
\setlength{\jot}{-0.2ex}
\begin{align}
\boldsymbol{q}_{\scriptscriptstyle\mathrm{WI}_j} =& 
\boldsymbol{q}_{\scriptscriptstyle\mathrm{WI}_i} \otimes 
\underset{\tilde{\boldsymbol{q}}_{\scriptscriptstyle\mathrm{I}_i \scriptscriptstyle\mathrm{I}_j}} 
{\underbrace{\int_{t \in \left[ i,j \right]} \boldsymbol{q}_{\scriptscriptstyle\mathrm{I}_i \scriptscriptstyle\mathrm{I}_t} \otimes 
\left[ \begin{matrix}
0 \\
\tilde{\boldsymbol{w}}_{\scriptscriptstyle\mathrm{I}_t} / 2
\end{matrix}\right]
dt}}\label{EQ: Rotational component of preintegration}, \\
\boldsymbol{v}_{\scriptscriptstyle\mathrm{WI}_j} =& 
\boldsymbol{v}_{\scriptscriptstyle\mathrm{WI}_i} - 
\boldsymbol{g}_{\scriptscriptstyle\mathrm{W}} \Delta t + 
\boldsymbol{R}_{ \scriptscriptstyle\mathrm{WI}_i} 
\underset{\tilde{\boldsymbol{\beta}}_{\scriptscriptstyle\mathrm{I}_i \scriptscriptstyle\mathrm{I}_j}}
{\underbrace{ \int_{t \in \left[ i,j \right]}\boldsymbol{R}_{\scriptscriptstyle\mathrm{I}_i 
\scriptscriptstyle\mathrm{I}_t}\tilde{\boldsymbol{a}}_{\scriptscriptstyle\mathrm{I}_t}dt}}
\label{EQ: Velocity part of preintegration}, \\ 
\boldsymbol{p}_{\scriptscriptstyle\mathrm{WI}_j} =& 
\boldsymbol{p}_{\scriptscriptstyle\mathrm{WI}_i} \!+\! 
\boldsymbol{v}_{\scriptscriptstyle\mathrm{WI}_i} \Delta t \!-\! 
\frac{1}{2} \boldsymbol{g}_{\scriptscriptstyle\mathrm{W}} \Delta t^2 \!+\! 
{ \boldsymbol{R}_{\scriptscriptstyle\mathrm{WI}_i} 
\underset{\tilde{\boldsymbol{\alpha}}_{\scriptscriptstyle\mathrm{I}_i \scriptscriptstyle\mathrm{I}_j}} 
{\underbrace{ \iint_{t \in \left[ i,j \right]}\boldsymbol{R}_{\scriptscriptstyle\mathrm{I}_i \scriptscriptstyle\mathrm{I}_t}
\tilde{\boldsymbol{a}}_{\scriptscriptstyle\mathrm{I}_t}dt^2}}}
\label{EQ: Translational component of preintegration}, 
\end{align}
\end{footnotesize}%
where $\Delta t$ is the time interval, and $\boldsymbol{g}_{\scriptscriptstyle\mathrm{W}}$ represents the gravity in the world frame $\underrightarrow{\mathcal{F}}_{\mathrm{W}}$. Rotation, velocity, and translation preintegrations are denoted by curly braces. For more details on the discrete-time implementation of preintegration and noise propagation, please refer to \cite{qin2018vins}.

\subsubsection{Extrinsic Rotation Recovery} Based on the rotational preintegration $\boldsymbol{q}_{\scriptscriptstyle\mathrm{I}_i \scriptscriptstyle\mathrm{I}_j}$, we formulate a constraint on the extrinsic rotation $\boldsymbol{q}_{\scriptscriptstyle\mathrm{MI}}$ using MoCap measurements between the two time instances: \par\vspace{-3pt}
\begin{small}
\begin{equation}
\scalebox{0.92}{$
\left( \tilde{\boldsymbol{q}}_{\scriptscriptstyle\mathrm{W} \scriptscriptstyle\mathrm{M}_i}^{\scriptscriptstyle{-1}} \otimes 
\tilde{\boldsymbol{q}}_{\scriptscriptstyle\mathrm{W} \scriptscriptstyle\mathrm{M}_j} \right) \otimes \boldsymbol{q}_{\scriptscriptstyle\mathrm{MI}} = \boldsymbol{q}_{\scriptscriptstyle\mathrm{MI}} \otimes \boldsymbol{q}_{\scriptscriptstyle\mathrm{I}_i \scriptscriptstyle\mathrm{I}_j}, 
$}
\end{equation}
\end{small}%
by expressing quaternion left and right multiplications in matrix form, denoted as $\left[ \cdot \right]_{\scriptscriptstyle{L}}$ and $\left[ \cdot \right]_{\scriptscriptstyle{R}}$, the constraint becomes a linear equation: \par\vspace{-3pt}
\begin{small}
\begin{equation}
\underset{\boldsymbol{Q}_{ij}} {\underbrace{ \left[ \left[ \tilde{\boldsymbol{q}}_{\scriptscriptstyle\mathrm{M}_i \scriptscriptstyle\mathrm{M}_j} \right]_{\mathrm{L}} - \left[ \boldsymbol{q}_{\scriptscriptstyle\mathrm{I}_i \scriptscriptstyle\mathrm{I}_j} \right]_{\mathrm{R}} \right] }}
 \boldsymbol{q}_{\scriptscriptstyle\mathrm{MI}} = \boldsymbol{0}.
 \label{EQ: Linear extrinsic rotation calibration}
\end{equation}
\end{small}%

To estimate $\boldsymbol{q}_{\scriptscriptstyle\mathrm{MI}}$, we stack multiple $\boldsymbol{Q}_{ij}$ from a motion sequence into an overdetermined linear equations and solve it using singular value decomposition (SVD). However, SVD is sensitive to noise and outliers, which are common in MoCap and IMU data. To improve robustness, we introduce a kernel function to enhance the signal-to-noise ratio. Based on screw theory \cite{pachtrachai2018chess}, the rotation angles $\theta_{\scriptscriptstyle\mathrm{M}_i \scriptscriptstyle\mathrm{M}_j}$ and $\theta_{\scriptscriptstyle\mathrm{I}_i \scriptscriptstyle\mathrm{I}_j}$ corresponding to non-identity relative rotations $\boldsymbol{q}_{\scriptscriptstyle\mathrm{M}_i \scriptscriptstyle\mathrm{M}_j}$ and $\boldsymbol{q}_{\scriptscriptstyle\mathrm{I}_i \scriptscriptstyle\mathrm{I}_j}$, are positive and equal in the absence of noise. 
This property allows us to define a weight function for $\boldsymbol{Q}_{ij}$ under noisy measurements: \par\vspace{-3pt}
\begin{small}
\begin{equation}
K_{ij} = \mathrm{exp}\left( \mu \left( 1- \frac{\mathrm{max} \left( \theta_{\scriptscriptstyle\mathrm{M}_i \scriptscriptstyle\mathrm{M}_j},\theta_{\scriptscriptstyle\mathrm{I}_i \scriptscriptstyle\mathrm{I}_j} \right)} {\mathrm{min} \left( \theta_{\scriptscriptstyle\mathrm{M}_i \scriptscriptstyle\mathrm{M}_j},\theta_{\scriptscriptstyle\mathrm{I}_i \scriptscriptstyle\mathrm{I}_j} \right)} \right) \right), 
\label{EQ: Robust kernel based on rotational screw constraint}
\end{equation}
\end{small}%
where $\mu$ is an empirically set amplification factor (5 in our work). The robust kernel is then applied to the linear equations for estimating extrinsic rotation, yielding \par\vspace{-3pt}
\begin{small}
\begin{equation}
\left[ K_{12}\boldsymbol{Q}_{12}^\mathrm{\scriptscriptstyle{T}}, K_{23}\boldsymbol{Q}_{23}^\mathrm{\scriptscriptstyle{T}}, \dots, K_{ij}\boldsymbol{Q}_{ij}^\mathrm{\scriptscriptstyle{T}} \right]^\mathrm{\scriptscriptstyle{T}} \boldsymbol{q}_{\scriptscriptstyle\mathrm{MI}} = \boldsymbol{0}.
\label{EQ: Robustified extrinsic rotation calibration}
\end{equation}
\end{small}%

\subsubsection{Velocity, Gravity, and Extrinsic Translation Initialization} Further, using the extrinsic parameters $\boldsymbol{R}_{\scriptscriptstyle\mathrm{MI}}$, $\boldsymbol{p}_{\scriptscriptstyle\mathrm{MI}}$ and MoCap measurement $\tilde{\boldsymbol{T}}_{\scriptscriptstyle\mathrm{WM}_i}$, we derive the following constraints: \par\vspace{-8pt}
\begin{small}
\begin{align}
\boldsymbol{R}_{\scriptscriptstyle\mathrm{WI}_i} =& \tilde{\boldsymbol{R}}_{\scriptscriptstyle\mathrm{WM}_i} \boldsymbol{R}_{\scriptscriptstyle\mathrm{MI}}, \\ 
\boldsymbol{p}_{\scriptscriptstyle\mathrm{WI}_i} =& \tilde{\boldsymbol{R}}_{\scriptscriptstyle\mathrm{WM}_i} \boldsymbol{p}_{\scriptscriptstyle\mathrm{MI}} + \tilde{\boldsymbol{p}}_{\scriptscriptstyle\mathrm{WM}_i}.
\end{align}
\end{small}%

Hence, the preintegration of velocity and translation in \eqref{EQ: Velocity part of preintegration} and \eqref{EQ: Translational component of preintegration} can be reformulated as \par\vspace{-5pt}
\begin{small}
\begin{equation}
\boldsymbol{H}_{ij}
\left[
\begin{matrix}
\boldsymbol{v}_{\scriptscriptstyle\mathrm{WI}_i} \\
\boldsymbol{v}_{\scriptscriptstyle\mathrm{WI}_j} \\
\boldsymbol{g}_{\scriptscriptstyle\mathrm{W}} \\
\boldsymbol{p}_{\scriptscriptstyle\mathrm{MI}} 
\end{matrix}
\right] = 
\left[ 
\begin{matrix}
\tilde{\boldsymbol{R}}_{\scriptscriptstyle\mathrm{WM}_i} \boldsymbol{R}_{\scriptscriptstyle\mathrm{MI}} \tilde{\boldsymbol{\alpha}}_{\scriptscriptstyle\mathrm{I}_i \scriptscriptstyle\mathrm{I}_j} + \tilde{\boldsymbol{p}}_{\scriptscriptstyle\mathrm{WM}_i} - 
\tilde{\boldsymbol{p}}_{\scriptscriptstyle\mathrm{WM}_j} \\
\tilde{\boldsymbol{R}}_{\scriptscriptstyle\mathrm{WM}_i} \boldsymbol{R}_{\scriptscriptstyle\mathrm{MI}} \tilde{\boldsymbol{\beta}}_{\scriptscriptstyle\mathrm{I}_i \scriptscriptstyle\mathrm{I}_j}
\end{matrix} 
\right], 
\label{EQ: Linear Initialization}
\end{equation}
\end{small}%
where 
\begin{small}
\begin{equation}
\boldsymbol{H}_{ij} = 
\left[ 
\begin{matrix}
-\boldsymbol{I}\Delta t & \boldsymbol{0} & \frac{1}{2} \Delta t^2  & -\tilde{\boldsymbol{R}}_{\scriptscriptstyle\mathrm{WM}_i} + 
\tilde{\boldsymbol{R}}_{\scriptscriptstyle\mathrm{WM}_j} \\
-\boldsymbol{I} & \boldsymbol{I} & \Delta t & \boldsymbol{0}
\end{matrix}
\right]. 
\end{equation}
\end{small}%

For consecutive time instances $t_{\mathrm{I}_{\scriptscriptstyle{i}}}$ to $t_{\mathrm{I}_{\scriptscriptstyle{j}}}$, we derive a nonhomogeneous linear equation involving velocity, gravity vector, and extrinsic translation. Similarly, by stacking constraints from continuous motion, the initialization is also reduced to a linear least squares problem  $\boldsymbol{Ax} = \boldsymbol{b}$, with a closed-form solution $\boldsymbol{x} = \left( \boldsymbol{A}^{\mathrm{\scriptscriptstyle{T}}} \boldsymbol{A} \right)^{\scriptscriptstyle{-1}} \boldsymbol{A}^{\mathrm{\scriptscriptstyle{T}}} \boldsymbol{b}$. To enhance robustness, we adopt a strategy similar to that in \eqref{EQ: Robust kernel based on rotational screw constraint}, but employ a screw constraint with translation along the principal axis as the kernel.

We perform initialization within the random sample consensus (RANSAC) framework to reject outliers, using the previously introduced rotational and translational constraints from screw theory as inlier metrics. Once the extrinsic and gravity alignment states defined in \eqref{EQ: Spatiotemporal extrinsic states} and \eqref{EQ: Gravity alignment states} are initialized, the MoCap poses are propagated to estimate the initial inertial states described in \eqref{EQ: Inertial states}. The IMU biases are initialized to zero, given their typically small magnitudes.

\subsection{Nonlinear Optimization} \label{SEC: Nonlinear Optimization}

\begin{figure}[t]
\centering
\includegraphics[width=0.8\linewidth]{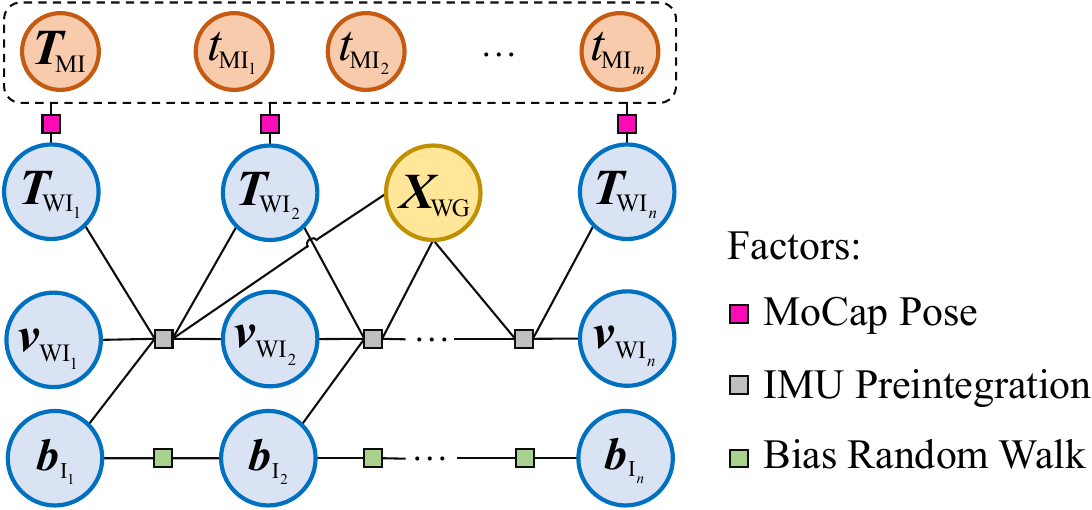}
\caption{Factor graph representation of the nonlinear optimization objective. Circles represent the states to be estimated, as defined in \eqref{EQ: Inertial states} -- \eqref{EQ: Gravity alignment states}, and squares represent the sensor factors.}
\label{FIG: Nonlinear factor graph}
\vspace{-13pt}
\end{figure}

After initialization, we refine the states through joint optimization using MoCap and IMU measurements within a MLE framework. The problem is formulated as a factor graph, as shown in Fig. \ref{FIG: Nonlinear factor graph}, and is solved by minimizing the negative log-likelihood function: \par\vspace{-5pt}
\begin{small}
\begin{align}
\underset{\boldsymbol{X}}{\mathrm{argmin}} \bigg( \;
{\textstyle \sum_{k=1}^{n}} & \left\| 
\boldsymbol{r}_{\scriptscriptstyle{\mathrm{M}_k}} 
\left( \tilde{\boldsymbol{T}}_{\scriptscriptstyle\mathrm{WM}_{k}}, 
\boldsymbol{T}_{\scriptscriptstyle\mathrm{WI}_k},  
\boldsymbol{X}_{\scriptscriptstyle\mathrm{MI}} \right) \right\|^2 
_{\Sigma_{\mathrm{M}_k}} \nonumber \\
+ {\textstyle \sum_{k=1}^{n-1}} & \left\| 
\boldsymbol{r}_{\scriptscriptstyle{\mathrm{I}_k}} 
\left( \tilde{\boldsymbol{P}}_{\scriptscriptstyle\mathrm{I}_k \scriptscriptstyle\mathrm{I}_{k+1}}, 
\boldsymbol{X}_{\scriptscriptstyle\mathrm{WI}_k}, 
\boldsymbol{X}_{\scriptscriptstyle\mathrm{WI}_{k+1}}, 
\boldsymbol{X}_{\scriptscriptstyle\mathrm{WG}} \right) \right\|^2 
_{\Sigma_{\mathrm{I}_k}} \nonumber \\
+ {\textstyle \sum_{k=1}^{n-1}} & \left\| 
\boldsymbol{b}_{\scriptscriptstyle{a,\mathrm{I}_{k+1}}} -
\boldsymbol{b}_{\scriptscriptstyle{a,\mathrm{I}_{k}}} \right\|^2 
_{\Sigma_{\boldsymbol{b}_{\scriptscriptstyle{a}}}}  \label{EQ: Loss function} \\
+ {\textstyle \sum_{k=1}^{n-1}} & \left\| 
\boldsymbol{b}_{\scriptscriptstyle{g,\mathrm{I}_{k+1}}} -
\boldsymbol{b}_{\scriptscriptstyle{g,\mathrm{I}_{k}}} \right\|^2 
_{\Sigma_{\boldsymbol{b}_{\scriptscriptstyle{g}}}} \nonumber
\bigg),
\end{align}
\end{small}%
where $\tilde{\boldsymbol{T}}_{\scriptscriptstyle\mathrm{WM}_{k}}$ and $\tilde{\boldsymbol{P}}_{\scriptscriptstyle\mathrm{I}_k \scriptscriptstyle\mathrm{I}_{k+1}}$ denote the MoCap pose and IMU preintegration measurements, respectively, and $\Sigma$ represents the covariances. The details of the sensor factors are presented in the remainder of this section.

\subsubsection{IMU factor} Based on the preintegration results in \eqref{EQ: Rotational component of preintegration}-\eqref{EQ: Translational component of preintegration}, we derive the 9-dimensional IMU residual $\boldsymbol{r}_{\scriptscriptstyle{\mathrm{I}_k}}$ for the consecutive inertial states $\boldsymbol{X}_{\scriptscriptstyle\mathrm{WI}_k}$ and $\boldsymbol{X}_{\scriptscriptstyle\mathrm{WI}_{k+1}}$ as \par\vspace{-5pt}
\begin{footnotesize}
\renewcommand{\arraystretch}{1.1}
\begin{equation}
\begin{aligned}
&\boldsymbol{r}_{\scriptscriptstyle{\mathrm{I}_k}} = \left[ \boldsymbol{r}_{{\scriptscriptstyle\boldsymbol{p,}}{\scriptscriptstyle{\mathrm{I}_k}}}^\mathrm{\scriptscriptstyle{T}}, \boldsymbol{r}_{{\scriptscriptstyle\boldsymbol{v,}}{\scriptscriptstyle{\mathrm{I}_k}}}^\mathrm{\scriptscriptstyle{T}}, \boldsymbol{r}_{{\scriptscriptstyle\boldsymbol{q,}}{\scriptscriptstyle{\mathrm{I_k}}}}^\mathrm{\scriptscriptstyle{T}} \right]^\mathrm{\scriptscriptstyle{T}} = \\ 
&\left[ 
\begin{matrix}
\boldsymbol{R}_{\scriptscriptstyle\mathrm{WI}_k}^{\scriptscriptstyle{-1}} \!
\left( \boldsymbol{p}_{\scriptscriptstyle\mathrm{WI}_{k+1}} \!\!-\! 
\boldsymbol{p}_{\scriptscriptstyle\mathrm{WI}_k} \!\!-\!
\boldsymbol{v}_{\scriptscriptstyle\mathrm{WI}_k} \Delta t \!+\!
\frac{1}{2} \boldsymbol{g}_{\scriptscriptstyle\mathrm{W}} \Delta t^2 \right) \!-\!
\tilde{\boldsymbol{\alpha}}_{\scriptscriptstyle\mathrm{I}_k \scriptscriptstyle\mathrm{I}_{k+1}} \\
\boldsymbol{R}_{\scriptscriptstyle\mathrm{WI}_k}^{\scriptscriptstyle{-1}}
\left( \boldsymbol{v}_{\scriptscriptstyle\mathrm{WI}_{k+1}} -
\boldsymbol{v}_{\scriptscriptstyle\mathrm{WI}_k} +
\boldsymbol{g}_{\scriptscriptstyle\mathrm{W}} \Delta t \right) -
\tilde{\boldsymbol{\beta}}_{\scriptscriptstyle\mathrm{I}_k \scriptscriptstyle\mathrm{I}_{k+1}} \\
2\left[ \tilde{\boldsymbol{q}}_{\scriptscriptstyle\mathrm{I}_k \scriptscriptstyle\mathrm{I}_{k+1}}^{\scriptscriptstyle{-1}} \otimes
\left( \boldsymbol{q}_{\scriptscriptstyle\mathrm{WI}_k}^{\scriptscriptstyle{-1}} \otimes
\boldsymbol{q}_{\scriptscriptstyle\mathrm{WI}_{k+1}} \right) \right]_{\mathrm{vec}} 
\end{matrix} 
\right],
\end{aligned}
\end{equation}
\end{footnotesize}%
where $\left[ \cdot \right]_{\mathrm{vec}}$ denotes the imaginary part of the quaternion, quantifying the rotation error. The vector $\boldsymbol{g}_{\scriptscriptstyle\mathrm{W}}$ represents gravity in the frame $\underrightarrow{\mathcal{F}}_{\mathrm{W}}$, and its Jacobian with respect to the gravity alignment state $\big[ \theta_{\scriptscriptstyle\mathrm{WG}}^{\scriptscriptstyle\mathrm{roll}}, \theta_{\scriptscriptstyle\mathrm{WG}}^{\scriptscriptstyle\mathrm{pitch}} \big]^\mathrm{\scriptscriptstyle{T}}$ can be obtained through the chain rule. The preintegration error terms robustly constrain consecutive inertial states over short intervals, reducing the effect of high-frequency MoCap jitter.

\subsubsection{MoCap factor}

For MoCap measurements, they provide globally drift-free constraints that effectively reduce the accumulation of errors caused by IMU integration. The MoCap residual $\boldsymbol{r}_{\scriptscriptstyle{\mathrm{M}}}$ corresponding to each inertial state $\boldsymbol{X}_{\scriptscriptstyle\mathrm{WI}_k}$ is defined as \par\vspace{-5pt}
\begin{small}
\begin{equation}
\renewcommand{\arraystretch}{1.5}
\boldsymbol{r}_{\scriptscriptstyle{\mathrm{M}_k}} = 
\left[ \begin{matrix}
\boldsymbol{r}_{{\scriptscriptstyle{\boldsymbol{p,}}}{\scriptscriptstyle{\mathrm{M}_k}}} \\
\boldsymbol{r}_{{\scriptscriptstyle{\boldsymbol{q,}}}{\scriptscriptstyle{\mathrm{M}_k}}}
\end{matrix} \right] =
\left[ \begin{matrix}
\boldsymbol{p}_{\scriptscriptstyle\mathrm{WI}_k} -
\boldsymbol{R}_{\scriptscriptstyle\mathrm{WI}_k} 
\boldsymbol{R}_{\scriptscriptstyle\mathrm{MI}}^{\scriptscriptstyle{-1}} 
\boldsymbol{p}_{\scriptscriptstyle\mathrm{MI}} - 
\tilde{\boldsymbol{p}}_{\scriptscriptstyle\mathrm{WM}_k} \\
2\left[ 
\tilde{\boldsymbol{q}}_{\scriptscriptstyle\mathrm{WM}_k}^{\scriptscriptstyle{-1}} \otimes 
\left(
\boldsymbol{q}_{\scriptscriptstyle\mathrm{WI}_k} \otimes 
\boldsymbol{q}_{\scriptscriptstyle\mathrm{MI}}^{\scriptscriptstyle{-1}} 
\right)
\right]_{\mathrm{vec}} 
\end{matrix} \right].
\end{equation}
\end{small}%

Since MoCap and IMU operate on different clocks, the timestamp of each inertial state is transformed to the MoCap clock via $\tau_{\scriptscriptstyle\mathrm{I}} = t_{\scriptscriptstyle\mathrm{I}} + t_{\scriptscriptstyle\mathrm{MI}}$, and the corresponding MoCap measurement $\tilde{\boldsymbol{T}}_{\scriptscriptstyle\mathrm{WM}_k}$ is then obtained by interpolation. Instead of the linear method used in \cite{geneva2020vicon2gt}, we adopt the cumulative cubic B-spline on the $SE\left(3\right)$ manifold proposed in \cite{sommer2020efficient} for temporal interpolation of MoCap data. Compared to the linear interpolation, cubic B-splines introduce a smoothness prior that ensures $C^2$ kinematic continuity, which helps reduce MoCap jitter. Furthermore, the step size for computing the Jacobian with respect to the time offset is determined by the interval between B-spline control points, effectively suppressing local gradient noise caused by high-frequency error.

Specifically, for a uniform cubic B-spline, the pose at any query time $\tau \in \left[ \tau_{\scriptscriptstyle\mathrm{M}_1}, \tau_{\scriptscriptstyle\mathrm{M}_2} \right)$ is determined by four consecutive control points \scalebox{0.9}{$\tilde{\boldsymbol{T}}_{\scriptscriptstyle\mathrm{WM}_i}$} (with $i \in \{0,1,2,3\}$) that are uniformly spaced in time, along with their corresponding MoCap timestamps $\tau_{\scriptscriptstyle\mathrm{M}_i}$. It can be expressed as \par\vspace{-5pt}
\begin{small}
\begin{equation}
\begin{aligned}
\tilde{\boldsymbol{T}}_{\scriptscriptstyle\mathrm{WM}}
\left( \tau \right) = &
\boldsymbol{\mathrm{exp}} \left( 
B_0 \left( \tau \right)
\boldsymbol{\mathrm{log}}
\left( \tilde{\boldsymbol{T}}_{\scriptscriptstyle\mathrm{WM}_0} 
\right) \right) \\
& {\textstyle \prod_{i=1}^{3}}
\boldsymbol{\mathrm{exp}}
\left( B_i \left( \tau \right)
\boldsymbol{\mathrm{log}}
\left( \tilde{\boldsymbol{T}}_{\scriptscriptstyle\mathrm{WM}_{i-1}}^{-1}
\tilde{\boldsymbol{T}}_{\scriptscriptstyle\mathrm{WM}_i} \right) 
\right), 
\label{EQ: Cubic B-spline}
\end{aligned}
\end{equation}
\end{small}%
where $\boldsymbol{\mathrm{exp}} \left( \cdot \right)$ denotes the exponential map from Lie algebra to Lie group, and $\boldsymbol{\mathrm{log}} \left( \cdot \right)$ represents the logarithm map, its inverse. The basis function $B_i \left( \tau \right)$ can be detailed as \par\vspace{-5pt}
\begin{small}
\begin{equation}
\begin{aligned}
\boldsymbol{B} \left( \tau \right) = &
\left [ \begin{array}
{@{\hspace{0em}}c@{\hspace{0.6em}}c@{\hspace{0.6em}}c@{\hspace{0.6em}}c@{\hspace{0em}}}
B_0 \left( \tau \right) & 
B_1 \left( \tau \right) & 
B_2 \left( \tau \right) &
B_3 \left( \tau \right) 
\end{array} \right ] \\ = &
\frac{1}{6} 
\left[ \begin{array}
{@{\hspace{0em}}c@{\hspace{0.6em}}c@{\hspace{0.6em}}c@{\hspace{0.6em}}c@{\hspace{0em}}}
1 & u \left( \tau \right) & 
u^2 \left( \tau \right) & 
u^3 \left( \tau \right)
\end{array} \right]
\left[ \begin{array}
{@{\hspace{0em}}c@{\hspace{0.6em}}c@{\hspace{0.6em}}c@{\hspace{0.6em}}c@{\hspace{0em}}}
 6 & 5 & 1 & 0 \\
 0 & 3 & 3 & 0 \\
 0 & -3 & 3 & 0 \\
 0 & 1 & -2 & 1
\end{array} \right], 
\end{aligned}
\end{equation}
\end{small}%
where $u\left( \tau \right) = \left( \tau - \tau_{\scriptscriptstyle\mathrm{M}_1} \right) / \left( \tau_{\scriptscriptstyle\mathrm{M}_2} - \tau_{\scriptscriptstyle\mathrm{M}_1} \right)$. 


In most existing studies, the time offset $t_{\scriptscriptstyle\mathrm{MI}}$ is treated as a constant parameter. However, local clocks without communication suffer from time scale errors that cause offset drift, especially during long-term operation. To address this, we model the time offset as a linear B-spline, which does not impose additional continuity constraints and is embedded within the cubic B-spline trajectory defined in \eqref{EQ: Cubic B-spline}. For any inertial timestamp $t_{\scriptscriptstyle\mathrm{I}_k}$, the corresponding time offset $t_{\scriptscriptstyle\mathrm{MI}_k}$ is calculated by weighting its two neighboring control points $t_{\scriptscriptstyle\mathrm{MI}_r}$ and $t_{\scriptscriptstyle\mathrm{MI}_s}$ as $t_{\scriptscriptstyle\mathrm{MI}_k} = \left( 1 - \lambda \right) t_{\scriptscriptstyle\mathrm{MI}_r} + \lambda t_{\scriptscriptstyle\mathrm{MI}_s}$, where $\lambda \in \left[0, 1\right]$ is the normalized weight factor. For each MoCap residual, the measurement is computed as follows: \par\vspace{-5pt}
\begin{footnotesize}
\begin{equation}
\begin{aligned}
\left[ \begin{matrix}
\tilde{\boldsymbol{R}}_{\scriptscriptstyle\mathrm{WM}_k} & 
\tilde{\boldsymbol{p}}_{\scriptscriptstyle\mathrm{WM}_k} \\
\boldsymbol{0} & 1
\end{matrix} \right] &=
\tilde{\boldsymbol{T}}_{\scriptscriptstyle\mathrm{WM}}
\left( t_{\scriptscriptstyle\mathrm{I}_k} + 
\left( 1 - \lambda \right) t_{\scriptscriptstyle\mathrm{MI}_r} + \lambda t_{\scriptscriptstyle\mathrm{MI}_s}
\right).
\end{aligned}
\end{equation}
\end{footnotesize}%

Therefore, we treat the time offset control points as states to be estimated, enabling variable time alignment. The Jacobian of each MoCap residual with respect to $\left[ t_{\scriptscriptstyle\mathrm{MI}_r},
t_{\scriptscriptstyle\mathrm{MI}_s}
\right]^\mathrm{\scriptscriptstyle{T}}$ is obtained via the chain rule as \par\vspace{-5pt}
\begin{footnotesize}
\begin{equation}
\begin{aligned}
\frac {\partial \boldsymbol{r}_{\scriptscriptstyle{\mathrm{M}_k}}}
{\partial \left[ t_{\scriptscriptstyle\mathrm{MI}_r},\!
t_{\scriptscriptstyle\mathrm{MI}_s}
\right]^\mathrm{\scriptscriptstyle{T}}} \!\! &= \!\!
\frac {\partial \bigl[ 
\boldsymbol{r}_{{\scriptscriptstyle{\boldsymbol{p,}}}{\scriptscriptstyle{\mathrm{M}_k}}}^\mathrm{\scriptscriptstyle{T}},\!
\boldsymbol{r}_{{\scriptscriptstyle{\boldsymbol{q,}}}{\scriptscriptstyle{\mathrm{M}_k}}}^\mathrm{\scriptscriptstyle{T}}
\bigr]^\mathrm{\scriptscriptstyle{T}}}
{\partial \bigl[ \tilde{\boldsymbol{p}}_{\scriptscriptstyle\mathrm{WM}_k}^\mathrm{\scriptscriptstyle{T}},\!
\tilde{\boldsymbol{q}}_{\scriptscriptstyle\mathrm{WM}_k}^\mathrm{\scriptscriptstyle{T}} 
\bigr]^\mathrm{\scriptscriptstyle{T}}}
\frac {\partial \bigl[ \tilde{\boldsymbol{p}}_{\scriptscriptstyle\mathrm{WM}_k}^\mathrm{\scriptscriptstyle{T}},\!
\tilde{\boldsymbol{q}}_{\scriptscriptstyle\mathrm{WM}_k}^\mathrm{\scriptscriptstyle{T}} 
\bigr]^\mathrm{\scriptscriptstyle{T}}} 
{\partial t_{\scriptscriptstyle\mathrm{MI}_k}}
\frac {\partial t_{\scriptscriptstyle\mathrm{MI}_k}}
{\partial \left[ t_{\scriptscriptstyle\mathrm{MI}_r},\!
t_{\scriptscriptstyle\mathrm{MI}_s}
\right]^\mathrm{\scriptscriptstyle{T}}}\! \\ &= 
\left[ \begin{matrix}
-\boldsymbol{I} & \boldsymbol{0}\\
\boldsymbol{0} & 
\boldsymbol{J}^{\boldsymbol{r}_{{\scriptscriptstyle{\boldsymbol{q,}}}{\scriptscriptstyle{\mathrm{M}_k}}}}_{
\tilde{\boldsymbol{q}}_{\scriptscriptstyle\mathrm{WM}_k}}
\end{matrix} \right]
\left[ \begin{matrix}
\dot{\boldsymbol{p}}_{\scriptscriptstyle\mathrm{WM}_k} \\
\dot{\boldsymbol{q}}_{\scriptscriptstyle\mathrm{WM}_k}^{\vee}
\end{matrix} \right] 
\left[ 
\begin{matrix}
1-\lambda & \lambda
\end{matrix} \right], 
\label{EQ: Time offset Jacobian}
\end{aligned}
\end{equation}
\end{footnotesize}%
where $\left( \cdot \right)^{\vee}$ denotes the vee map operator, and \par\vspace{-5pt}
\begin{small}
\begin{equation}
\boldsymbol{J}^{\boldsymbol{r}_{{\scriptscriptstyle{\boldsymbol{q,}}}{\scriptscriptstyle{\mathrm{M}_k}}}}_{
\tilde{\boldsymbol{q}}_{\scriptscriptstyle\mathrm{WM}_k}} = -
\left[ \begin{matrix}
\boldsymbol{0} & \boldsymbol{I}
\end{matrix} \right] 
\left[ \tilde{\boldsymbol{q}}_{\scriptscriptstyle\mathrm{WM}_k} \right]_{\mathrm{L}}
\left[ \boldsymbol{q}_{\scriptscriptstyle\mathrm{MI}} \otimes
\boldsymbol{q}_{\scriptscriptstyle\mathrm{WI}_k}^{\scriptscriptstyle{-1}} \right]_{\mathrm{R}} 
\left[ \begin{matrix}
\boldsymbol{0} \\
\boldsymbol{I}
\end{matrix} \right].
\end{equation}
\end{small}%

\subsubsection{Degenerate Identification} For spatiotemporal calibration, adequate motion excitation is critical to ensure accurate estimation. For any two MoCap residuals $\boldsymbol{r}_{\scriptscriptstyle{\mathrm{M}_k}}$ and $\boldsymbol{r}_{\scriptscriptstyle{\mathrm{M}_{k^{\prime}}}}$, the difference in their translational components is given by \par\vspace{-5pt}
\begin{small}
\begin{equation}
\begin{aligned}
\boldsymbol{r}&_{{\scriptscriptstyle{\boldsymbol{p,}}}{\scriptscriptstyle{\mathrm{M}_{k^{\prime}}}}} - 
\boldsymbol{r}_{{\scriptscriptstyle{\boldsymbol{p,}}}{\scriptscriptstyle{\mathrm{M}_k}}} = \\
&\boldsymbol{R}_{\scriptscriptstyle\mathrm{WI}_k}
\boldsymbol{p}_{{\scriptscriptstyle\mathrm{I}_k}{\scriptscriptstyle\mathrm{I}_{k^{\prime}}}} \!-\! \left( \boldsymbol{R}_{\scriptscriptstyle\mathrm{WI}_{k^{\prime}}} \!-\! \boldsymbol{R}_{\scriptscriptstyle\mathrm{WI}_k}\right)
\boldsymbol{R}_{\scriptscriptstyle\mathrm{MI}}^{\scriptscriptstyle{-1}} 
\boldsymbol{p}_{\scriptscriptstyle\mathrm{MI}} \!-\!
\tilde{\boldsymbol{R}}_{\scriptscriptstyle\mathrm{WM}_k}
\tilde{\boldsymbol{p}}_{{\scriptscriptstyle\mathrm{M}_k}{\scriptscriptstyle\mathrm{M}_{k'}}}.
\end{aligned}
\end{equation}
\end{small}%
When the motion degenerates toward pure translation, we have approximately  $\boldsymbol{R}_{\scriptscriptstyle\mathrm{WI}_k}
\boldsymbol{p}_{{\scriptscriptstyle\mathrm{I}_k}{\scriptscriptstyle\mathrm{I}_{k^{\prime}}}} = \tilde{\boldsymbol{R}}_{\scriptscriptstyle\mathrm{WM}_k}
\tilde{\boldsymbol{p}}_{{\scriptscriptstyle\mathrm{M}_k}{\scriptscriptstyle\mathrm{M}_{k'}}}$, $\boldsymbol{R}_{\scriptscriptstyle\mathrm{WI}_{k^{\prime}}} = \boldsymbol{R}_{\scriptscriptstyle\mathrm{WI}_k}$, indicating that the translational residuals become nearly linearly dependent, resulting in an ill-conditioned problem. As the motion further approaches a stationary state, the rotational residuals also lose observability. 

Therefore, sufficient rotation is essential to ensure well-constrained solution. We divide the MoCap data into consecutive time windows of size $w$. For each window $\left[ t_n, t_n+w \right]$, the maximum rotation angle $\theta_n$ is computed as \par\vspace{-5pt}
\begin{footnotesize}
\begin{equation}
\theta_n=\underset{\substack{t_i,t_j \in \left[ t_n, t_n+w \right]}}{\max} 
\arccos\Bigg(
\frac{\mathrm{trace} \big(
\tilde{\boldsymbol{R}}_{\scriptscriptstyle\mathrm{WM}}^{\scriptscriptstyle{-1}} \left( t_i \right)
\tilde{\boldsymbol{R}}_{\scriptscriptstyle\mathrm{WM}}\left( t_j \right)\big) \!-\! 1}{2}
\Bigg).
\end{equation}
\end{footnotesize}%
If $\theta_n < \varpi$, the window is considered degenerate. In this case, the MoCap factor constrains only the inertial states and does not provide gradients for the spatiotemporal calibration parameters, which helps to reduce the impact of noise. In our work, $w$ and $\varpi$ are empirically set to 5 s and 10$^\circ$, respectively.

Based on the MoCap and IMU factors described above, we solve the nonlinear least squares problem in \eqref{EQ: Loss function} with the Ceres solver \cite{agarwal2022ceressolver}. The IMU-centric high-precision GT trajectory can be obtained from the estimated states in \eqref{EQ: All states}.

\section{Experimental Results} \label{SEC: Experimental Results}

\begin{table}[!t]
\renewcommand{\arraystretch}{1.35}
\caption{Sensor Noise Parameters in the Self-Collected Dataset}
\label{Table: Sensor Noise}
\centering
\begin{tabular}{l|c}
\toprule[1.0pt] 
\multicolumn{1}{c|}{Noise type} & \multicolumn{1}{c}{Value} \\ 
\midrule
Accelerometer noise density ($m/\mathrm{s}^2/\sqrt\mathrm{Hz}$) & $5.2 \times 10^{-3}$ \\
Accelerometer random walk ($m/\mathrm{s}^3/\sqrt\mathrm{Hz}$)   & $1.0 \times 10^{-3}$ \\
Gyroscope noise density ($rad/\mathrm{s}/\sqrt\mathrm{Hz}$)     & $2.1 \times 10^{-4}$ \\
Gyroscope random walk ($rad/\mathrm{s}^2/\sqrt\mathrm{Hz}$)     & $1.3 \times 10^{-5}$ \\
MoCap translational noise density ($m/\sqrt\mathrm{Hz}$)        & $4.3 \times 10^{-5}$ \\
MoCap rotational noise density ($rad/\sqrt\mathrm{Hz}$)         & $1.7 \times 10^{-4}$ \\
\bottomrule[1.0pt]
\end{tabular}
\end{table}

This section presents three sets of experiments to validate our algorithm. Section \ref{SEC: Simulated Experiments} evaluates key modules and overall estimator performance on simulated data. Section \ref{SEC: Real-World Experiments} demonstrates the accuracy of the algorithm on both public and self-collected real-world datasets. Section \ref{SEC: SLAM Benchmarking Applications} demonstrates the application of our method on SLAM benchmarking with public datasets, emphasizing the crucial role of GT quality in error metric quantification. In addition to our proposed \textbf{MoCap2GT}, we evaluated three representative approaches for comparison:
\begin{enumerate}[label=\arabic*., itemsep=0.3ex]
    \item \textbf{HEC-Filter} \cite{furrer2018evaluation, shu2024spatiotemporal}: Hand-eye calibration with additional median filtering \cite{schubert2018tum} to suppress MoCap jitter.
    \item \textbf{Vicon2GT} \cite{geneva2020vicon2gt}: A batch estimator based on IMU pre-integration and linear interpolation of MoCap poses. 
    \item \textbf{Kalibr-M} \cite{rehder2016extending}: The MoCap branch of the Kalibr toolbox, performing joint optimization based on MoCap and IMU data within a continuous-time framework.
\end{enumerate}
The self-collected datasets were acquired using a Vicon Viro 2.2 MoCap system and a TDK InvenSense ICM-42688-P IMU. Key noise parameters, obtained from static calibration of both sensors, are summarized in Table \ref{Table: Sensor Noise}.

\subsection{Simulated Experiments} \label{SEC: Simulated Experiments}

The simulation test was first conducted to enable a controlled evaluation of the proposed method. The trajectory estimated by ORB-SLAM3 \cite{campos2021orb} on the TUM-VI \cite{schubert2018tum} room1 sequence served as the basis trajectory for our simulator, which was used to generate IMU measurements at 500 Hz and MoCap data at 100 Hz as inputs to the algorithm. Following standard practice, IMU readings were corrupted by white noise and random walk bias, while MoCap measurements were affected by white noise. The default sensor noise parameters were set to match those of the devices used in our self-collected dataset, as shown in Table \ref{Table: Sensor Noise}. Additionally, the time offset drift between the MoCap and IMU was modeled as a deviation in clock scale, with the default noise set to 2 ms per minute. To evaluate algorithm performance under varying noise levels, default noise parameters were scaled from 0.2$\times$ to 2$\times$ in steps of 0.2, resulting in 10 noise levels in our tests. When the noise of one sensor was varied, the others were kept at default.

\begin{figure}[t]
\centering
\includegraphics[width=\linewidth]{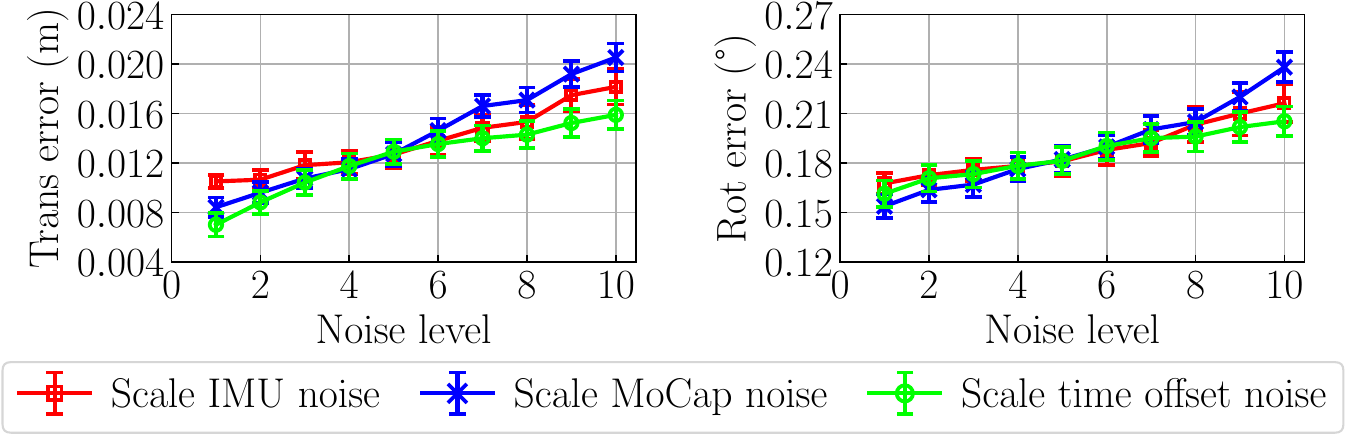}
\caption{Performance of the linear initializer on the simulated dataset across different noise types and levels. Subplots show the mean and standard deviation of translation and rotation errors from repeated trials. }
\label{FIG: Performance of the linear initializer on the simulator}
\vspace{-13pt}
\end{figure}

\subsubsection{Initializer assessment} The performance of our linear initializer was evaluated under varying noise types and levels. Fig. \ref{FIG: Performance of the linear initializer on the simulator} shows the translation and rotation errors of the spatial extrinsic parameters estimated by the initializer, which served as key metrics during the initialization stage. By employing robust kernels derived from screw theory and an iterative linear least squares solver, the initializer demonstrates strong robustness against various increasing noise levels. Notably, even at the highest noise level, it maintains mean translation and rotation errors below 0.02 m and 0.24$^\circ$, respectively, providing reliable initial estimates that facilitate the convergence of the nonlinear optimizer.

\begin{figure}[t]
\centering
\includegraphics[width=0.9\linewidth]{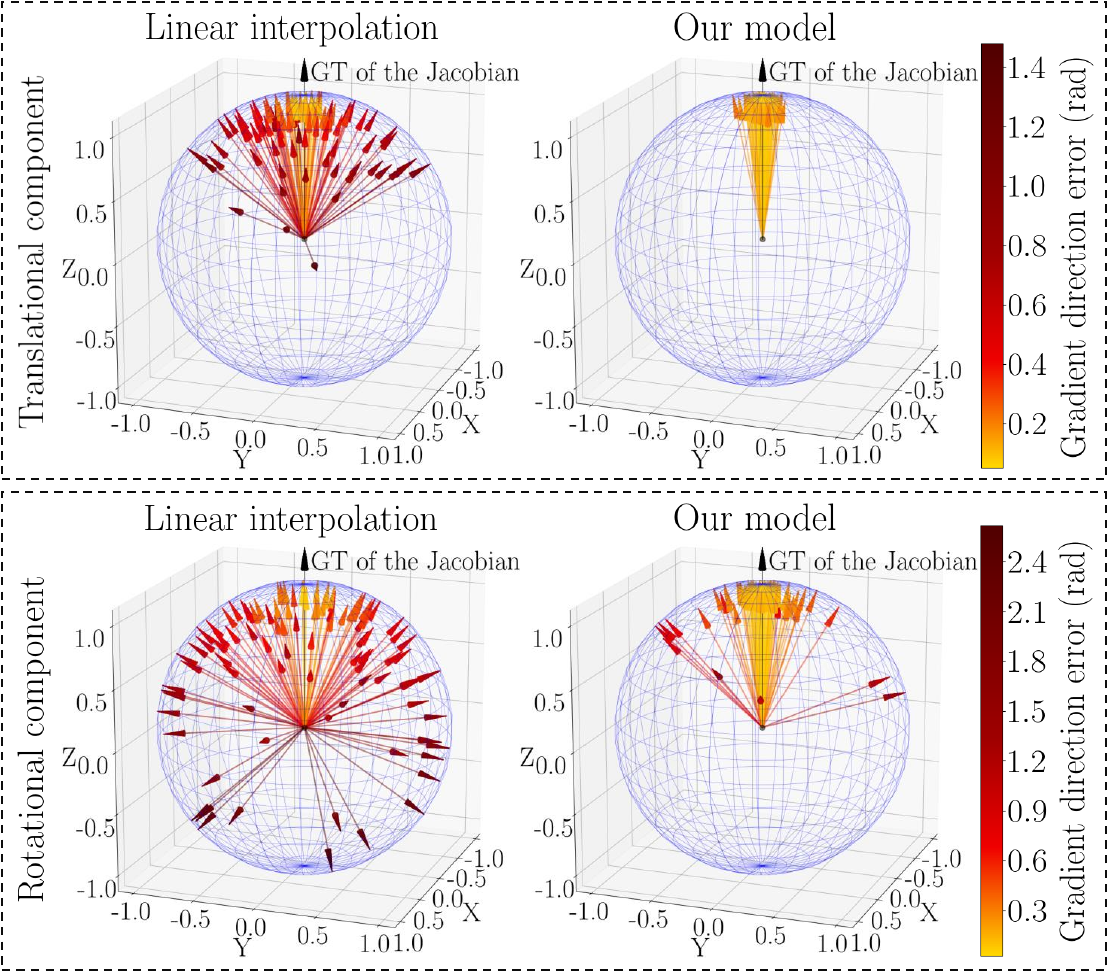}
\caption{Comparison of Jacobians corresponding to MoCap residuals with respect to time offset for different MoCap models. Translational and rotational components are normalized and visualized on a unit sphere to illustrate directional errors. }
\label{FIG: Comparison of the time offset Jacobian}
\vspace{-13pt}
\end{figure}

\subsubsection{MoCap model verification} The effectiveness of the MoCap measurement model introduced in Section \ref{SEC: Nonlinear Optimization} was validated through a comparison with the linear interpolation method adopted by Vicon2GT \cite{geneva2020vicon2gt}. The accuracy of the Jacobians corresponding to MoCap residuals with respect to the time offset served as the evaluation metric, as it was critical for fusing MoCap and IMU data in nonlinear optimization. Simulated data with fixed time offset derivatives under default noise were generated, and the Jacobians computed from both methods were assessed based on their directional errors relative to the GT. Since both the translational and rotational components of the Jacobians were three-dimensional vectors, they were visualized in Euclidean space, with directional errors indicated by color variations as shown in Fig. \ref{FIG: Comparison of the time offset Jacobian}. Compared to the linear interpolation method, the proposed model achieves smaller directional errors, especially for rotational components. The results demonstrate the robustness of the model to local high-frequency noise and the principled modeling of MoCap data.

\begin{figure}[t]
\centering
\includegraphics[width=0.9\linewidth]{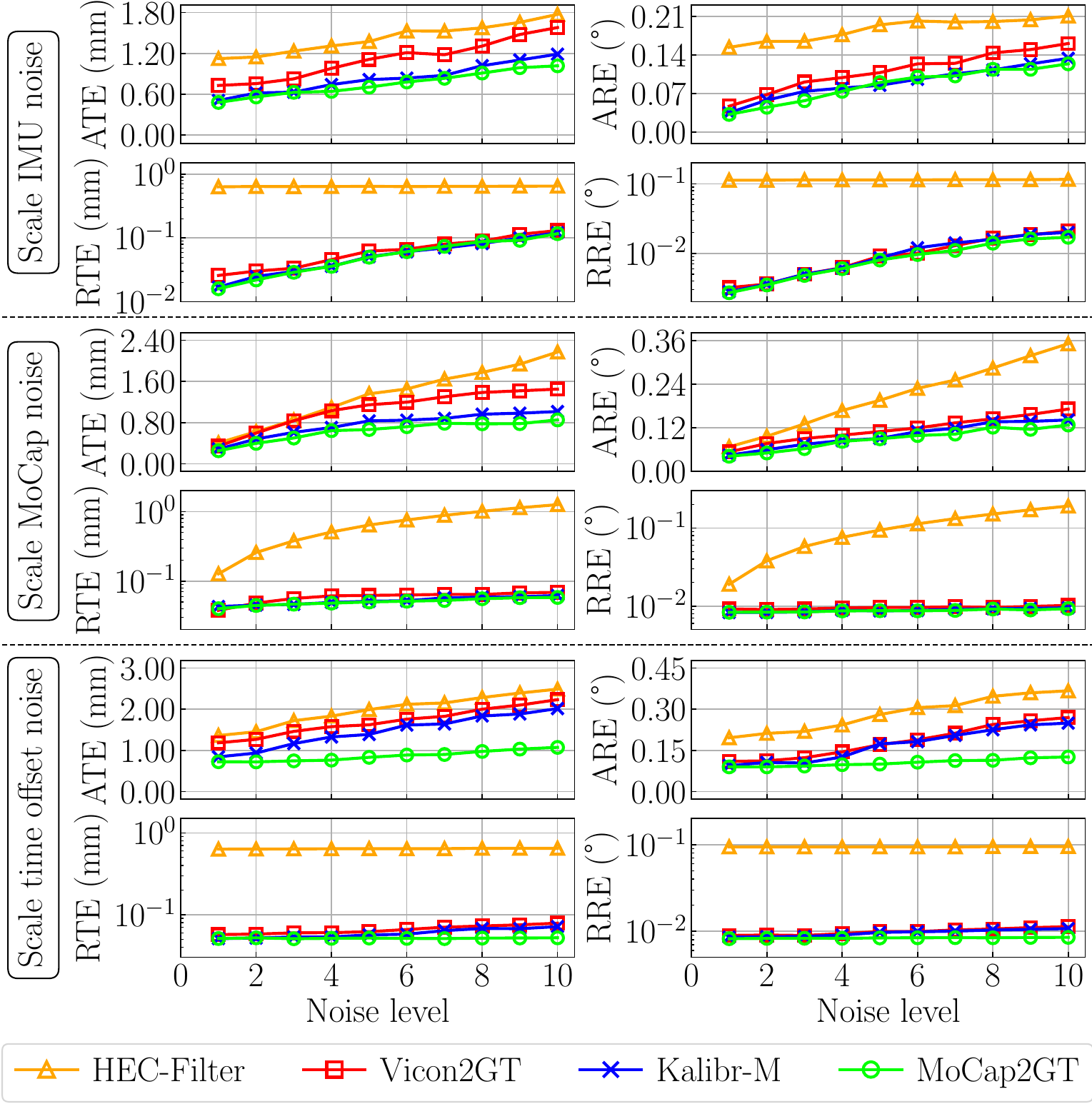}
\caption{Performance comparison of different methods on the simulated dataset. Results are shown for trajectory errors under varying noise types and levels.}
\label{FIG: Comparison of different GT estimators}
\end{figure}

\subsubsection{Overall estimator evaluation} An overall evaluation of the proposed GT estimator was conducted in comparison with existing methods. Since the compared algorithms do not support time-varying time offset estimation, clock drift was set to zero when assessing the impact of MoCap and IMU noise. All estimated trajectories were uniformly resampled to 50 Hz, and the four trajectory error metrics introduced in Section \ref{SEC: Introduction} were computed with their respective root mean square errors (RMSE) to evaluate performance. The comparison results are presented in Fig. \ref{FIG: Comparison of different GT estimators}. Among all methods, MoCap2GT achieves the highest accuracy across all metrics. HEC-Filter fails to effectively handle MoCap jitter, resulting in large relative errors that also contribute to increased absolute errors. Vicon2GT and Kalibr-M leverage IMU measurements to compensate for MoCap jitter, achieving RTE and RRE comparable to those of MoCap2GT. However, their ATE and ARE remain higher, primarily due to errors in spatiotemporal calibration. As noise levels increase, all methods exhibit degraded performance. Notably, when time offset drift occurs, methods assuming a fixed offset suffer a significant drop in accuracy, whereas MoCap2GT maintains low errors. Overall, MoCap2GT achieves the best performance and remains robust under varying conditions.

\subsection{Real-World Experiments} \label{SEC: Real-World Experiments}

\begin{table}[t]
\caption{Comparison of Spatial Calibration Results for Different Methods. Reported Values Are Mean $\pm$ Standard Deviation over Repeated Experiments, with Best Results in Bold}
\label{TAB: Comparison of Spatial Calibration}
\centering
\setlength{\tabcolsep}{5pt}
\begin{tabular}{c|c|cc}
\toprule[1.0pt]
Dataset                         & Method          & Trans (cm)      & Rot ($^\circ$)  \\ \midrule
\multirow{4}{*}{TUM-VI}         & HEC-Filter      & 7.352$\pm$0.109 & 1.291$\pm$0.079 \\
                                & Vicon2GT        & 7.228$\pm$0.121 & 1.322$\pm$0.090 \\
                                & Kalibr-M        & 7.321$\pm$0.095 & 1.219$\pm$\textbf{0.066} \\
                                & MoCap2GT (ours) & 7.293$\pm$\textbf{0.086} & 1.224$\pm$0.068 \\ \midrule
\multirow{4}{*}{Self-collected} & HEC-Filter      & 4.641$\pm$0.097 & 2.994$\pm$0.073 \\
                                & Vicon2GT        & 4.587$\pm$0.105 & 3.015$\pm$0.069 \\
                                & Kalibr-M        & 4.653$\pm$0.084 & 3.072$\pm$0.055 \\
                                & MoCap2GT (ours) & 4.660$\pm$\textbf{0.072} & 3.063$\pm$\textbf{0.049} \\
\bottomrule[1.0pt]
\end{tabular}
\vspace{-13pt}
\end{table}

In addition to simulations, we compared different methods on both public and self-collected real-world datasets. Due to the presence of errors in the official GT of public datasets, there is a lack of reliable reference for directly evaluating trajectory accuracy. Therefore, we first focused on calibration quality, which is critical to overall algorithm performance. Multiple calibrations were performed on the same device to assess the consistency of different methods. The test data included IMU calibration sequences from the TUM-VI dataset and data collected using our own MoCap and IMU setup. We analyzed the translational and rotational components of the spatial extrinsics in the calibration, with the mean and standard deviation of results from different methods reported in Table \ref{TAB: Comparison of Spatial Calibration}. The mean values obtained by different methods showed some variations, with the maximum discrepancy reaching 0.13 cm/0.10$^\circ$ on the TUM-VI dataset and 0.07 cm/0.08$^\circ$ on our dataset. Among these methods, MoCap2GT demonstrates the highest calibration consistency, as evidenced by its smallest standard deviation.

\begin{figure}[t]
\centering
\includegraphics[width=0.9\linewidth]{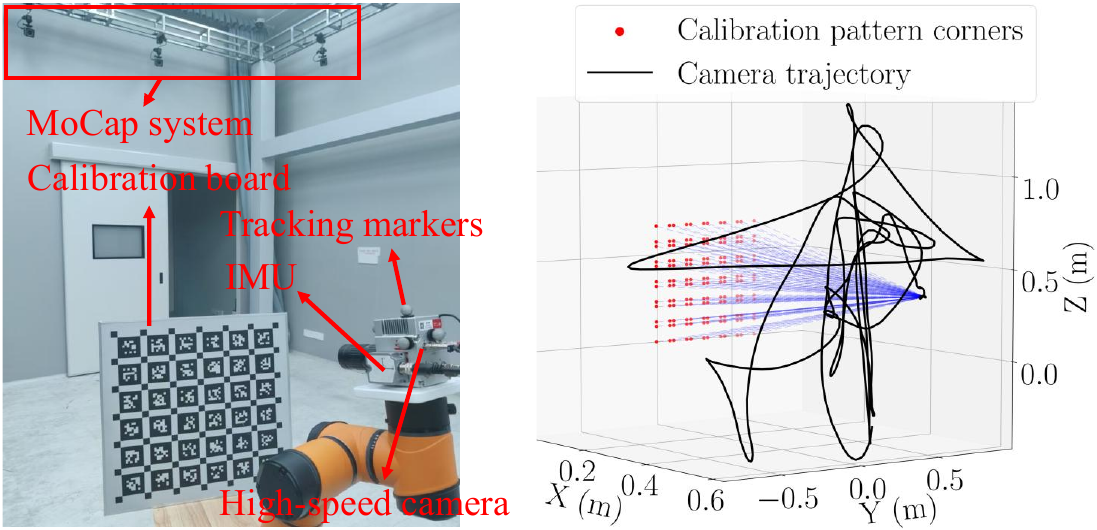}
\caption{Experimental setup (left) and illustration of trajectory reconstruction using global bundle adjustment (right) in real-world experiments.}
\label{FIG: Hardware setup in real-world experiments}
\end{figure}

To directly evaluate the trajectory accuracy of the proposed method, we built an experimental system as shown on the left side of Fig. \ref{FIG: Hardware setup in real-world experiments}. A low-distortion, high-speed 2K camera was used to capture a calibration board at close range, and the motion trajectory was recovered using global bundle adjustment (BA), as illustrated on the right side of Fig. \ref{FIG: Hardware setup in real-world experiments}. Based on precise corner detections from the calibration board, the BA achieved a reprojection error of less than 0.1 pixels. At a viewing distance of 1 m, this theoretically corresponds to a pose error of 0.05 mm and 0.003$^\circ$, providing sufficiently accurate reference trajectories for evaluating our algorithm. The MoCap system and IMU were rigidly attached to the high-speed camera and collected data simultaneously. The estimated trajectories were resampled to 50 Hz, aligned with the reference trajectories, and then compared to compute trajectory errors. It is worth noting that this setup has a certain limitation: to ensure the calibration board remains within the field of view of the high-speed camera during motion, testing can only be performed within a limited range.

\begin{figure}[t]
\centering
\includegraphics[width=0.95\linewidth]{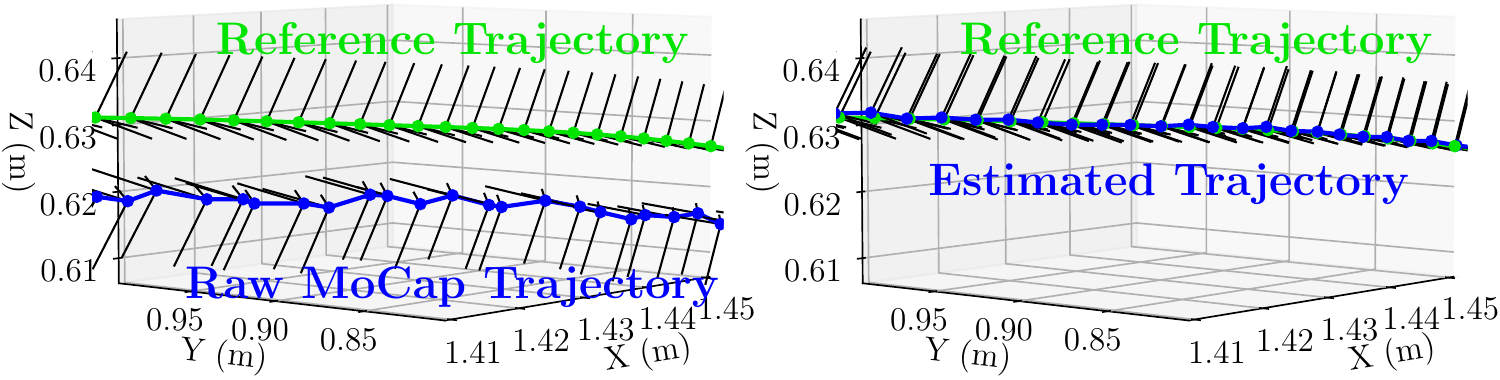}
\caption{Visual comparison of the raw MoCap data (left) and the result estimated by MoCap2GT (right), both shown with the reference trajectory. To enhance visual clarity, the magnitude of jitter is amplified by a factor of 5. }
\label{FIG: Visual comparison of the trajectory}
\vspace{-13pt}
\end{figure}

\begin{table}[!t]
\caption{Comparison of Different Methods on Real-World Datasets. Results That Meet the Targets* Are Highlighted in Green and Best Results Are Bolded}
\label{TAB: Comparison on Real-World Datasets}
\centering
\renewcommand{\arraystretch}{0.95}
\setlength{\tabcolsep}{4pt}
\begin{tabular}{c| c| c c c c}
\toprule[1.0pt]
Scenario                & Algorithm       & \begin{tabular}[c]{@{}c@{}}ATE\\ (mm)\end{tabular} 
                                          & \begin{tabular}[c]{@{}c@{}}ARE\\ ($^\circ$)\end{tabular} 
                                          & \begin{tabular}[c]{@{}c@{}}RTE\\ (mm)\end{tabular} 
                                          & \begin{tabular}[c]{@{}c@{}}RRE\\ ($^\circ$)\end{tabular} \\ \midrule 
\multirow{4}{*}{\shortstack{Sufficient \\ Motion}} & HEC-Filter      & 2.251 & 0.257 & 0.605 & 0.115 \\
                        & Vicon2GT        & 2.178 & 0.242 & \colorbox{green!30}{0.217} & \colorbox{green!30}{0.015} \\
                        & Kalibr-M        & \colorbox{green!30}{1.534} & 0.207 & \colorbox{green!30}{0.183} & \colorbox{green!30}{\textbf{0.013}}  \\
                        & MoCap2GT (ours) & \colorbox{green!30}{\textbf{1.466}} & \colorbox{green!30}{\textbf{0.178}} & \colorbox{green!30}{\textbf{0.177}} & \colorbox{green!30}{\textbf{0.013}} \\ \midrule
\multirow{4}{*}{\shortstack{Motion \\ Degradation}} & HEC-Filter      & 3.262 & 0.291 & 0.610 & 0.109 \\
                        & Vicon2GT        & 3.543 & 0.277 & \colorbox{green!30}{0.229} & \colorbox{green!30}{0.017} \\
                        & Kalibr-M        & 2.135 & 0.235 & \colorbox{green!30}{0.198} & \colorbox{green!30}{0.015} \\
                        & MoCap2GT (ours) & \colorbox{green!30}{\textbf{1.681}}  & \colorbox{green!30}{\textbf{0.173}}  & \colorbox{green!30}{\textbf{0.182}} & \colorbox{green!30}{\textbf{0.013}} \\
\bottomrule[1.0pt]
\multicolumn{6}{@{}l@{}}{\scriptsize{$^*$Targets outlined in Section \ref{SEC: Motivation}: ATE/ARE$<$2 mm/0.2$^\circ$, RTE/RRE$<$0.4 mm/0.02$^\circ$.}}
\end{tabular}
\end{table}

Fig. \ref{FIG: Visual comparison of the trajectory} provides a visual comparison of the data before and after processing by our estimator. The raw MoCap data, shown on the left, exhibits noticeable jitter and misalignment. In contrast, the estimated result is effectively compensated and well aligned with the reference trajectory. The quantitative comparison results are presented in Table \ref{TAB: Comparison on Real-World Datasets}, covering two scenarios: one with sufficient motion excitation and one with degraded motion. Consistent with simulation results, HEC-Filter produces the largest errors, as it does not fuse additional measurements to compensate for high-frequency MoCap noise. Vicon2GT and Kalibr-M effectively suppress the jitter but lack precise spatiotemporal calibration to ensure low ATE and ARE. MoCap2GT provides the most accurate results and robustly handles degraded scenarios. Notably, using consumer-grade IMU and MoCap, only our method fully meets the accuracy targets outlined in Section \ref{SEC: Motivation}, as highlighted in green.

\subsection{SLAM Benchmarking Applications} \label{SEC: SLAM Benchmarking Applications}

\begin{table}[!t]
\caption{Benchmarking Results of ORB-SLAM3 on the TUM-VI and EuRoC Datasets with Respect to Different GT References. Abnormal Results Are Highlighted in Red.}
\label{TAB: SLAM Benmarking Results}
\centering
\renewcommand{\arraystretch}{0.95}
\setlength{\tabcolsep}{4pt}
\begin{tabular}{c| c| c c c c}
\toprule[1.0pt]
Dataset                & GT source       & \begin{tabular}[c]{@{}c@{}}ATE\\ (mm)\end{tabular} 
                                          & \begin{tabular}[c]{@{}c@{}}ARE\\ ($^\circ$)\end{tabular} 
                                          & \begin{tabular}[c]{@{}c@{}}RTE\\ (mm)\end{tabular} 
                                          & \begin{tabular}[c]{@{}c@{}}RRE\\ ($^\circ$)\end{tabular} \\ \midrule 
\multirow{5}{*}{\shortstack{TUM-VI \\ room4}} & Official      & 19.082 & \colorbox{red!30}{0.602} & 5.943 & \colorbox{red!30}{0.362} \\
                        & HEC-Filter        & 19.321 & \colorbox{red!30}{0.613} & 5.995 & \colorbox{red!30}{0.342} \\
                        & Vicon2GT        & 18.962 & 0.534 & 5.321 & 0.071 \\
                        & Kalibr-M        & 18.152 & 0.543 & 5.291 & 0.067  \\
                        & MoCap2GT (ours) & 18.269 & 0.491 & 5.287 & 0.064 \\ \midrule
\multirow{5}{*}{\shortstack{TUM-VI \\ room5}} & Official      & 38.631 & \colorbox{red!30}{0.531} & 6.711 & \colorbox{red!30}{0.382} \\
                        & HEC-Filter        & 39.281 & \colorbox{red!30}{0.529} & 6.823 & \colorbox{red!30}{0.365} \\
                        & Vicon2GT        & \colorbox{red!30}{41.752} & \colorbox{red!30}{0.695} & 6.653 & 0.111 \\
                        & Kalibr-M        & 38.912 & 0.412 & 6.231 & 0.101  \\
                        & MoCap2GT (ours) & 38.505 & 0.423 & 6.187 & 0.098 \\ \midrule
\multirow{5}{*}{\shortstack{EuRoC \\ V101}} & Official      & \colorbox{red!30}{40.421} & \colorbox{red!30}{5.499} & 3.817 & 0.041 \\
                        & HEC-Filter        & 27.124 & 0.542 & \colorbox{red!30}{4.010} & \colorbox{red!30}{0.285} \\
                        & Vicon2GT        & 26.251 & 0.462 & 3.363 & 0.040 \\
                        & Kalibr-M        & 25.727 & 0.475 & 3.321 & 0.041  \\
                        & MoCap2GT (ours) & 25.631 & 0.451 & 3.329 & 0.039 \\ \midrule
\multirow{5}{*}{\shortstack{EuRoC \\ V203}} & Official      & 29.541 & \colorbox{red!30}{0.725} & 5.741 & 0.105 \\
                        & HEC-Filter        & 30.193 & 0.672 & 6.432 & \colorbox{red!30}{0.291} \\
                        & Vicon2GT        & 31.253 & 0.573 & 5.762 & 0.093 \\
                        & Kalibr-M        & 29.664 & \colorbox{red!30}{0.702} & 5.745 & 0.092  \\
                        & MoCap2GT (ours) & 29.178 & 0.584 & 5.739 & 0.093 \\
\bottomrule[1.0pt]
\end{tabular}
\vspace{-13pt}
\end{table}

Finally, we applied our method to practical SLAM benchmarking using the widely used TUM-VI \cite{schubert2018tum} and EuRoC \cite{burri2016euroc} datasets. In addition to the official GT trajectories, both datasets provide raw MoCap and IMU data, which enables re-computation of IMU-centric GT using different methods. We ran ORB-SLAM3 on both datasets and evaluated trajectory errors with respect to different GT references, with RMSE results for several sequences summarized in Table \ref{TAB: SLAM Benmarking Results}. 

In the TUM-VI dataset, benchmark results based on the official GT show large RRE and RTE due to ineffective compensation of MoCap jitter, which also increases ARE. While the official GT in EuRoC reduces MoCap noise through batch estimation, the accuracy of extrinsic calibration degrades in some sequences, particularly V101, causing high ATE and ARE. This problem has also been reported in several studies \cite{geneva2020vicon2gt, geneva2020openvins, shu2024spatiotemporal}. Moreover, other GT estimators exhibit reduced accuracy in some sequences, with affected results highlighted in red in the table. In contrast, MoCap2GT yields accurate results across all sequences. These findings further emphasize that insufficient GT accuracy remains a common challenge in SLAM benchmarking, particularly when quantifying rotation and inter-frame errors. Our method enables high-precision GT estimation and offers a promising solution to this problem.

\section{Conclusion} \label{SEC: Conclusion}

This paper presents MoCap2GT, a GT estimator based on the fusion of MoCap and IMU data for SLAM benchmarking. The proposed method revisits two key challenges in accurate trajectory estimation: spatiotemporal calibration and MoCap jitter mitigation, and provides effective solutions to recover high-precision GT trajectories. Detailed simulations and real-world experiments demonstrate that our approach outperforms existing methods and is of great significance for SLAM benchmarking. To our knowledge, this is the first method to meet the accuracy requirements in Section \ref{SEC: Motivation} using a consumer-grade IMU and MoCap system. Future work will extend the estimator to incorporate cameras, enabling a unified calibration framework. We also intend to release a visual-inertial dataset with high-precision GT and calibration results to support rigorous benchmarking and continued advancement in SLAM research.

\bibliographystyle{bib/IEEEtran}
\bibliography{bib/mybib}

\end{document}